\documentclass[conference]{IEEEtran}
\ifCLASSINFOpdf
   \usepackage[pdftex]{graphicx}
   \usepackage{subcaption}
\else
\fi
\usepackage{amsfonts}
\usepackage{amsmath}
\usepackage[noend]{algpseudocode}

\usepackage{multirow}

\usepackage{algorithm}
\usepackage{fancyhdr}
\usepackage{color}
\usepackage{array}
\usepackage{tabularx}
\usepackage{eso-pic}
\AddToShipoutPicture*{\small \sffamily\raisebox{1.2cm}{\hspace{1.8cm}This is a preprint of the paper accepted to ECMR 2021}}

\begin{document}

%
\title{Augmenting GRIPS with Heuristic Sampling for Planning Feasible Trajectories of a Car-Like Robot
}

\author{\IEEEauthorblockN{Brian Angulo}
\IEEEauthorblockA{Moscow Institute of Physics \\ and Technology}

\and
\IEEEauthorblockN{Ivan Radionov}
\IEEEauthorblockA{LLC ``Integrant''}

\and
\IEEEauthorblockN{Konstantin Yakovlev}
\IEEEauthorblockA{Federal Research Center ``Computer Science \\and Control'' RAS, Moscow Institute of \\ Physics and Technology}
}



%


\maketitle

\begin{abstract}

Kinodynamic motion planning for non-holomonic mobile robots is a challenging problem that is lacking a universal solution. One of the computationally efficient ways to solve it is to build a geometric path first and then transform this path into a kinematically feasible one. Gradient-informed Path Smoothing (GRIPS)~\cite{GRIPS} is a recently introduced method for such transformation. GRIPS iteratively deforms the path and adds/deletes the waypoints while trying to connect each consecutive pair of them via the provided steering function that respects the kinematic constraints. The algorithm is relatively fast but, unfortunately, does not provide any guarantees that it will succeed. In practice, it often fails to produce feasible trajectories for car-like robots with large turning radius. In this work, we introduce a range of modifications that are aimed at increasing the success rate of GRIPS for car-like robots. The main enhancement is adding an additional step that heuristically samples the waypoints along the bottleneck parts of the geometric paths (such as sharp turns). The results of the experimental evaluation provide a clear evidence that the success rate of the suggested algorithm is up to 40\% higher compared to the original GRIPS and hits the bar of 90\%, while its runtime is lower.

\end{abstract}

\IEEEpeerreviewmaketitle

\section{Introduction}

Motion planning is one of the fundamental problems in robotics and AI. Researchers from the AI community typically abstract away from the real-world complications and are concentrated on creating involved search and learning techniques to solve the most basic version of the problem, i.e. geometric planning, when the resultant path is composed of the straight-line segments. On the other hand, the robotic researchers wish to construct not just geometric paths, but the trajectories that respect the kinematic constraints of the robot under consideration -- the so-called kinodynamic planning.

Two conventional approaches to kinodynamic planning are the lattice-based and the sampling-based planning. Lattice-based planners assume that a set of the  kinematically feasible motion primitives is defined beforehand and is given to the planner as the input~\cite{MotionPrimitives}. They search over the lattice of the properly aligned primitives to find the trajectory. Sampling-based planners, such as RRT~\cite{RRT} or RRT*~\cite{RRT*}, grow a search tree by sampling states in the robot's configuration space and use a steering function to connect two states respecting the kinematic constraints of the robot. Thus they can be thought of as creating the motion primitives online (while planning). Both of these approaches are computationally demanding.

\begin{figure}[t]
    \centering
    \begin{subfigure}[b]{0.2\textwidth}
        \includegraphics[width=\textwidth]{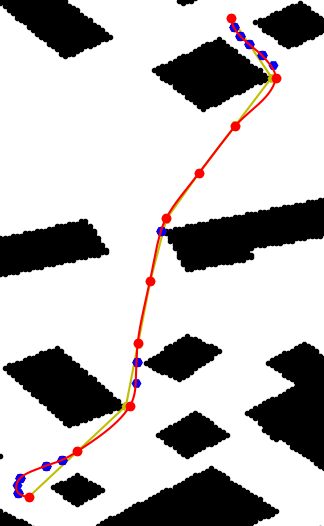}
        \label{fig:local}
    \end{subfigure}
    ~ 
    \begin{subfigure}[b]{0.2\textwidth}
        \includegraphics[width=\textwidth]{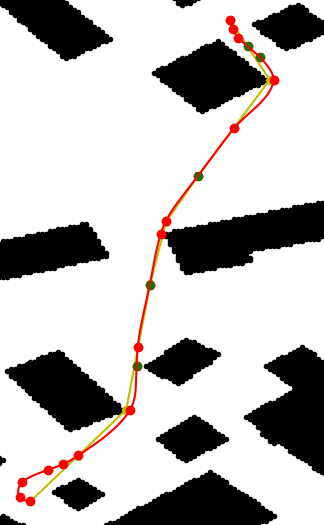}
        \label{fig:irremovable}
    \end{subfigure}
    \caption{Illustration of the GRIPS post-smoothing algorithm. \textbf{Blue} points represent the added states at the local minima after one iteration of the gradient descent. \textbf{Green} points are the irremovable states. \textbf{Red} curve is the trajectory that connects the subsequent states via the steering function. \textbf{Yellow} poly-line shows the geometric path.}\label{fig:PhasesGRIPS}
\end{figure}

One of the prominent approaches to create a fast kinodynamic planner is to decompose the problem into the two phases. First, a geometric path, i.e. a sequence of the straight-line segments, is built. This can be done very fast by invoking any state-of-the-art heuristic search planner or RRT-planner that operates in $(x, y)$ state space. Second, this path is transformed into a kinematically feasible one by a separate algorithm. GRIPS~\cite{GRIPS} is one of such algorithms, introduced recently and being able to post-process geometric paths very fast\footnote{In general GRIPS allows to transform any path, not only a geometric one, but in this work, we study GRIPS only in combination with the geometric planners.}. It operates by relocating some waypoints of the path while adding/deleting the other ones -- see Figure~\ref{fig:PhasesGRIPS}. Connecting the waypoints is done via the given steering function that respects the kinematic constraints of the robot.

Unfortunately, GRIPS does not provide any guarantees that the given geometric path will be successfully transformed into a kinodynamic trajectory. In practice, it may not work well for car-like robots that have a large turning radius. The reason is that the algorithm operates only with the waypoints of the provided path. To this end we suggest enhancing GRIPS by adding an optional exploration procedure that is invoked when certain conditions are met, i.e. when the algorithm is struggling to connect the waypoints with the kinematically-feasible motions. While, this procedure might be implemented in various ways, in this work we implement it as deterministic sampling of states guided by a heuristic rule.

Additionally, we suggest to modify one of the transformation steps GRIPS is performing, i.e. identifying and skipping the redundant states, to make the algorithm faster while not compromising the quality of the solution.

We evaluate the suggested modified GRIPS empirically and compare it with the original algorithm showing that the success rate (in planning for a car-like robot) notably increases from 50\% to 90\% in a wide range of realistic scenarios. Moreover, the computation time is lower by up to 80\% while the resultant length of the trajectories was almost the same.

The rest of the paper is organized as follows. An overview of the related works is presented in Section II. The problem is formally stated in Section III. In Section IV we describe the proposed method. Experimental setup and results are presented in Sections V.

\section{Related work}

The problem of kinodynamic planning (for car like robots) is well studied in the literature and various approaches based on (combination of) graph-based, sampling based, optimization and other techniques are known, see~\cite{gonzalez2015review} for review.

In seminal work of Likhachev and Ferguson~\cite{MotionPrimitives} a lattice-based planner that accounts for the car kinematic constraints was introduced. Other variants of the lattice-based planners are described in ~\cite{rufli2010design, ziegler2009spatiotemporal}. Contrary to these algorithms the suggested method does not construct a lattice in the high-dimensional space to search for a feasible plan.

Perhaps, somewhat more popular within the robotic community approach to kinodynamic planning is sampling based planning. One way to account for robot's dynamics within such type of planning is to propagate the randomly sampled control inputs~\cite{hsu2002randomized}. Another is to sample in the robot's state space and attempt to connect states via the steering function~\cite{hwan2011anytime, webb2013kinodynamic, BIT*SQP}. Method described in this work also relies on the steering function but it is not a sampling-based planner.

Noteworthy, approaches that combine lattice-based and sampling-based approaches exist~\cite{RRT*MotionPrimitives}. 

Perhaps, the most similar in spirit method to the one presented in this work is Theta*-RRT~\cite{THETA*RRT}. This method also uses a geometric path as the initial guess for the kinodynamic trajectory and also relies on the steering function. However, unlike our algorithm Theta*-RRT is a full-scale sampling based planner that biases the sampling in favor of the states lying in the proximity of the geometric path. In this work we compare with Theta*-RRT and report similar success rates while the runtime of our algorithm is notably lower.




Finally, it is also worth to mention another class of approaches that also use a reference path as the initial guess and then apply (involved) optimization techniques to transform this guess to the final solution -- see~\cite{OPTIMAL} for example. Unlike these approaches we rely on a rather simplistic and computationally efficient steering function to solve the two-boundary value problems arising while post-smoothing a geometric path.  



\section{Problem Statement}

We consider a car-like robot that has to navigate from the start state to the goal one in a static and fully-observable environment. The task is to find the corresponding trajectory that respects the kinematic constraints of the robot. 

The state-space of the robot is induced by the tuples $x(t) = (x, y, \theta)$, where $(x, y)$ are the coordinates of the robot, $\theta$ is the orientation. The control space is induced by the tuples $u(t) = (v, \gamma)$, where $v$ is the linear velocity and $\gamma$ is the steering angle. The robot's dynamic is given as:

\begin{align}\label{eq:diffEquationsRobot}
    &\dot{x} = v cos(\theta)\nonumber\\
    &\dot{y} = v sin(\theta)\\ 
    &\dot{\theta} = \frac{v}{L} \tan(\gamma), \nonumber
\end{align}
where $L$ is the wheel-base of the robot. 


The free configuration space of the robot is $\mathcal{X}_{free}$ and the task is to find the controls (as functions of time) that move the robot from its start state to the goal one while always staying in $\mathcal{X}_{free}$. The cost of the trajectory is its length. In this work we are not aimed at solving the problem optimally but low-cost solutions are desirable. 

\section{Method}

We build a geometric path first and then transform this path into a kinematically feasible trajectory. Any geometric planner can be used for the first step. At the second step, we use a modification of the previously-introduced GRIPS~\cite{GRIPS} algorithm. GRIPS accepts as input a geometric path $\sigma$ and a steering function \textit{St}. The latter is used to generate motion primitives that connects intermediate states, $\sigma[i] = (x, y, \theta)$, of the path. We first explain how the steering function works and then proceed to our enhanced GRIPS algorithm.

\subsection{Steering function}

Given the two distinct states, $\sigma[i]$, $\sigma[j]$, the task of the steering function is to generate a sequence of controls $(v, \gamma)$ to move the robot from the one state to the other ensuring the kinematic feasibility of the trajectory. 

In this work we will be using the steering function described in \cite{Corke2013Robotics} that implements the linear control law given by

\begin{align}\label{eq:ControlLaw}
    &v = k_{\rho} \rho \\
    &\omega = k_{\alpha} \alpha + k_{\beta} \beta \nonumber.
\end{align}

\begin{figure}[t]
    \centering
    \includegraphics[width = 0.6 \linewidth]{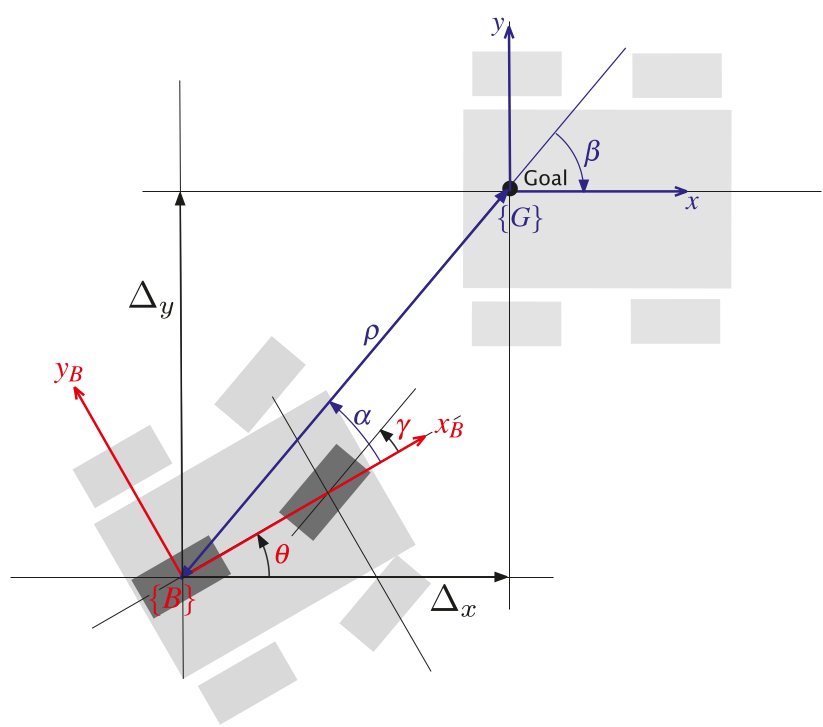}
    \caption{A car-like robot moving from one state to the other. Here {B} is the vehicle frame and {G} is the goal frame (image from \cite{Corke2013Robotics})}.
    \label{fig:CarLike}
\end{figure}

Here $\rho$ is the Euclidean distance between the states $\sigma[i]$, $\sigma[j]$, denoted as $B$ and $G$ on Figure~\ref{fig:CarLike}. $\alpha$ is the angle of the goal vector with respect to the frame $B$, $\beta$ is the angle of the goal vector with respect to the $G$. $\omega$ is the rotational velocity. The latter can be transformed into the steering angle as follows: $\gamma = \arctan \dfrac{\omega \cdot L}{v}$.

Eq.~\ref{eq:ControlLaw} defines a reactive controller. To get the resultant trajectory we simulate application of these controls. Thus the trajectory is represented as a sequence of states $\mathbf{tr}=(tr[0], ..., tr[n])$. In practice, one needs to define the simulation step $dt$ and the limit on the number of simulations to apply such steering. In our experiments we set them to 0.1s and 300 respectively. We considered the steering to be successful if the target state is reached with some tolerance (otherwise steering would be very troublesome). We account for the fact that steering may lead not exactly to the target state while planning.

\subsection{Modified Gradient Informed Path Smoothing with Heuristic Sampling}

As our modified version of GRIPS is similar to the original one, we will briefly describe the latter first.

\subsubsection{GRIPS background}
GRIPS algorithm consists of two major phases: \textit{MoveAndInsertStates} and \textit{PruneStates} -- see Alg.~\ref{alg:GRIPS}. Their pseudo-code is given in the original paper~\cite{GRIPS}. Here we will give a conceptual overview.

The first phase, \textit{MoveAndInsertStates}, moves the states of the geometric path $\sigma$ (except the start and the goal one) away from the obstacles and adds some extra states. In moving states GRIPS relies on the distance map $D$ which stores in each element the number that tells how far this element is from the nearest obstacle. GRIPS uses the gradient descent $\nabla D$ along the distance map to move away each $\sigma[i]$. The number of times each state in $\sigma$ is moved, $K$, is defined by the user as well as the gradient descent step $\eta$ and the discount factor $\gamma$ (not to be confused with the steering angle).

After the states of the geometric path $\sigma$ are moved away from the obstacles the same procedure \textit{inserts} additional states to $\sigma$. To do that the steering function is invoked:  $St(\sigma[i]$, $\sigma[i + 1])$, and each state of the resultant trajectory, $tr[k]$, is analyzed. If the distance from $tr[k]$ to the nearest obstacle is smaller than the one of the $tr[k-1]$ or $tr[k+1]$ and the distance between $tr[k]$ and $\sigma[i]$ (and $\sigma[i + 1]$) exceeds the user-predefined threshold $d_{\min}$ than $tr[k]$ is added to $\sigma$. These added states are shown in blue in Figure~\ref{fig:PhasesGRIPS}.

The second major phase of GRIPS, \emph{PruneStates}, removes some states from the current path $\sigma$ while connecting the others with the steering function $St$ (thus the resultant path respects the kinematic constraints). This procedure subsequently processes the states $\sigma[i-1], \sigma[i]$, $\sigma[i+1]$ and tries to create a shortcut manoeuvre from $\sigma[i-1]$ to $\sigma[i+1]$ (via the steering function). If this maneuver leads to a collision with the obstacle, the state $\sigma[i]$ is marked as irremovable. After the whole path has been processed and all the irremovable states have been identified the algorithm builds a directed acyclic graph (DAG) between each pair of them. The nodes in this graph are the intermediate states on the path, that were not marked as irremovable before (red points in Figure~\ref{fig:PhasesGRIPS}), the edges -- are the transitions between the states. After the DAG is built a shortest path on it is sought that connect the source irremovable state to the target one. Intuitively, the idea of \emph{PruneStates} is twofold. On the one hand, GRIPS tries to minimize the number of states in the path, on the other hand, it wants to minimize the cost (length) of the trajectory.

During both phases, the heading of each state $\sigma[i]$ is constantly updated by taking the average of the headings of $\sigma[i-1]$ and $\sigma[i + 1]$ (these are the headings of the incoming line from the predecessor and the heading of the outgoing line to the successor). This change allows for a smoother transition of the robot from one state to another.

\begin{algorithm}[t]
\caption{GRIPS}
\label{alg:GRIPS}
\begin{algorithmic}[1]
    \Function{GRIPS}{$\sigma$, $St$}
    \State $\sigma \gets$  \Call{MoveAndInsertStates}{$\sigma$, $St$}
    \State{\color{blue} $\sigma \gets$ \Call{PruneStates}{$\sigma$, $St$}}
    \State \Return $\sigma$
    \EndFunction
\end{algorithmic}
\end{algorithm}

\begin{algorithm}[t]
\caption{Prune States}
\label{alg:NewSecondPhase}
\begin{algorithmic}[1]
    \Function{PruneStates}{$\sigma$, $St$}
    \State $i \gets 0$, $N \gets |\sigma|$, $\sigma[z] \gets \sigma[0]$ 
    \State $\sigma' \gets$ \Call{Append}{$\sigma[0]$}
    \State $\diamond$ $H$ $\gets$ the horizon to reach the next states
    \While{$i < N - 1$}
        \If{$\neg$ \Call{Collides}{$St(\sigma[z], \sigma[i+1])$}}
            \State $i \gets$ \Call{SkipStates}{$\sigma[z]$, $i$, $St$}
        \Else
            \If{$H > 1$}
                \State $i \gets$ \Call{ReachNextStates}{$\sigma[z]$, $i$, $H$, $St$}
            \EndIf
            \If{$H < 1$  or \Call{ReachNextStates}{} failed}
                \State$\sigma[k] \gets$ \Call{ExtraStates}{$\sigma[z]$, $\sigma[i+1]$, $St$}
                \If{we couldn't provide a transition}
                    \State \Return $\emptyset$
                \Else
                    \State $i \gets i + 1$, $\sigma' \gets$ \Call{Append}{$\sigma[k]$}
                \EndIf
            \EndIf
        \EndIf
        \State $\sigma' \gets$ APPEND($\sigma[i]$)
        \State $\sigma[z] \gets$ the state that finishes the last motion
    \EndWhile
    \State \Return $\sigma'$
    \EndFunction
\end{algorithmic}
\end{algorithm}

\subsubsection{Modifications of GRIPS}

\begin{figure}[t]
    \centering
    \includegraphics[width=0.5\textwidth]{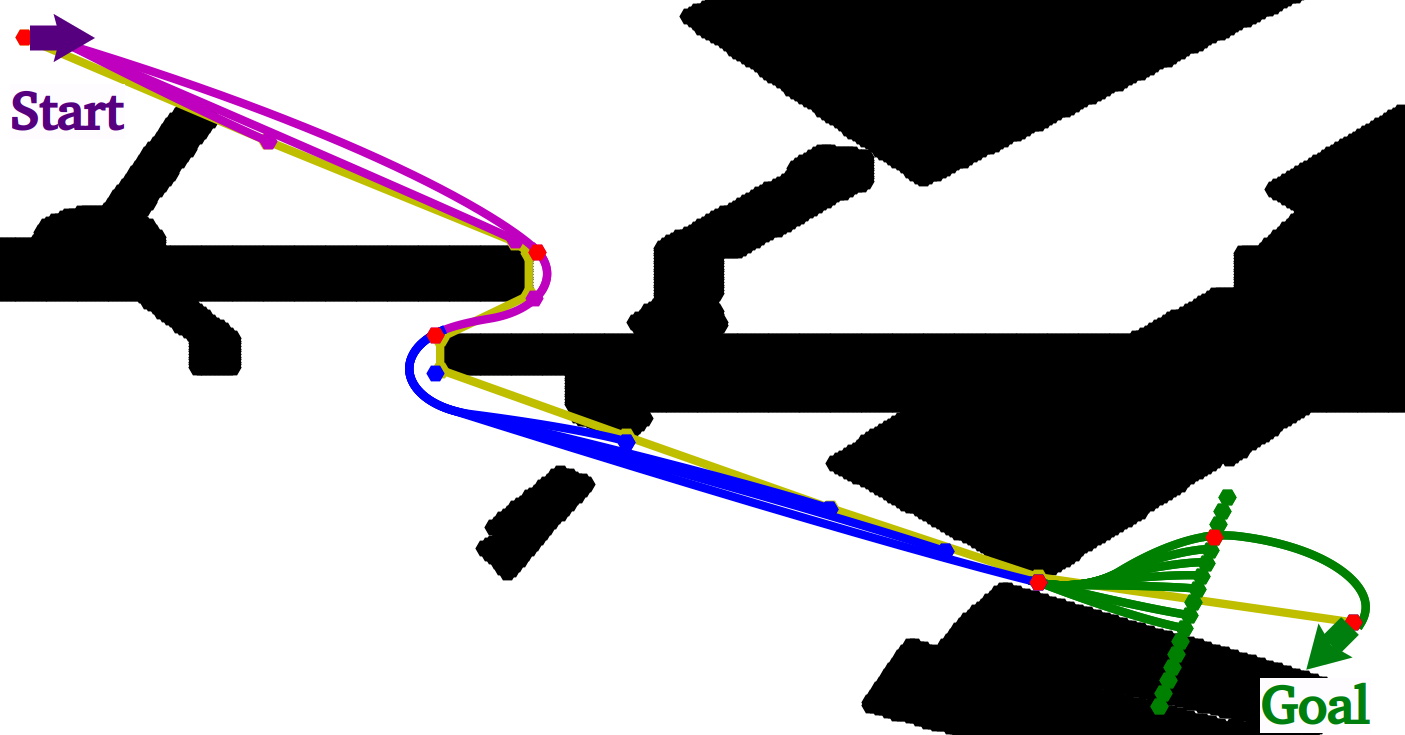}
    \caption{\textbf{Yellow} polyline shows the geometric path. \textbf{Purple}, \textbf{Blue}, \textbf{Green} curves and points represent the \textit{SkipStates}, \textit{ReachNextStates} and \textit{ExtraStates} respectively. \textbf{Red} points denote the states of the final path.}
    \label{fig:SecondPhase}
\end{figure}

Our version of GRIPS also uses the two major phases as in original GRIPS. While the first phase is the same, the second one is completely different. It builds on the three procedures: \textit{SkipStates}, \textit{ReachNextStates} and \textit{AddExtraState}. The idea is to greedily connect the current state to the next ones lying on the geometric path via the steering function. When it is impossible to connect two nearby states we try to reach the subsequent states in a predefined horizon. Finally, we employ adding extra states \emph{not lying on a geometric path} when steering through only the path states is impossible. We now explain these procedures using the running example depicted on Figure~\ref{fig:SecondPhase} and then describe the pseudo-code that implements them.

The first procedure is aimed as constructing a part of the trajectory starting at the current state of a geometric path and ending at one of the subsequent states by greedily \textit{skipping} the intermediate states. Skipping is performed until steering from the source state to a subsequent one fails. This might happen either when a motion primitive produced by a steering function collides with the obstacles or when the steering function fails to produce controls that transfer the robot to the target state. On Figure~\ref{fig:SecondPhase} the skipping connections are shown in purple. One can note that out of the six first states of the trajectory (start including) three were skipped and the resultant part of the trajectory now consists out of two segments (the red dots depict their endpoints).

The second procedure is invoked when two subsequent states can not be connected via the steering function. In the considered example this happens when transitioning from the endpoint state of the purple segment of the trajectory. In such, case we try to \textit{reach} the subsequent states of the geometric path. This process is performed either until it fails or until the reaching horizon $H$, specified by the user is exceeded. On Figure~\ref{fig:SecondPhase} the reaching connections are shown in blue. As one can see, utilizing this procedure allowed one to build a single prolonged segment of the trajectory almost up to the goal.

The third procedure is invoked when two subsequent states, denote them $A$ and $B$, can not be directly connected by the steering function and the \textit{ReachNextStates} procedure fails as well. This procedure adds extra states that do not belong to the geometric path to the workspace and tries to reach the target state $B$ via one of them. In general, a variety of ways to implement adding extra states can be proposed. In this work, we suggest the following approach that can be viewed as deterministic heuristic sampling. We place additional states (shown as green dots on Figure~\ref{fig:SecondPhase}) along the line segment $l$ that intersects the current segment of the geometric path, i.e. $AB$, perpendicularly and is positioned at its center. The length of $l$ is determined by a user-specified parameter $M$. The states are evenly distributed along $l$ with the distance between each two added states being $step$. The heading of each state coincides with the heading of the $AB$ segment of the geometric path.  

The pseudo-code of the modified \textit{PruneStates} procedure which is composed of the above-described algorithmic blocks is presented in Alg.~\ref{alg:NewSecondPhase}. $\sigma[z]$ denotes the anchor state, i.e. the state from which we are trying to build the next segment of the kinodynamically feasible trajectory. Initially, it is the start of the geometric path. In the main loop we progress towards the goal by finding the applicable motion from $\sigma[z]$ to one of the next waypoints of the given geometric path (Lines 6-10) or to one on the heuristically sampled waypoints (Lines 11-12).


Recall that steering from one state to the other may not exactly reach the target state. That is why we consistently update $\sigma[z]$ at Line 18 with the state that actually was the last on the trajectory produced by the steering function.

\begin{algorithm}[t]
\caption{Add Extra States}
\label{alg:ExtraStates}
\begin{algorithmic}[1]
    \Function{ExtraStates}{$\sigma[i]$, $\sigma[j]$, $St$}
    \State $\diamond$ $\beta$ is the average of headings of $\sigma[i]$ and $\sigma[j]$   
    \State $\diamond$ $\alpha$ is the slope of the segment $\langle \sigma[i]$, $\sigma[j] \rangle$
    \State $\diamond$ $(x, y)$ is the midpoint of the this segment
    \For{$d = -M, \ldots, M$}
        \State $x' \gets$ $x + d \cdot \cos\beta$
        \State $y' \gets$ $y + d \cdot \sin\beta$
        \If{$(x', y')$ is free}
            \State $\sigma[k]$ $\gets$ $(x', y', \alpha)$
            \If{$\neg$ \Call{Collides}{$St(\sigma[i], \sigma[k])$} $\wedge$ \State $\neg$ \Call{Collides}{$St(\sigma[k], \sigma[j])$}}
                \State \Return $\sigma[k]$
            \EndIf
        \EndIf
        \State $d = d + step$ 
    \EndFor 
    \State \Return $\emptyset$
    \EndFunction
\end{algorithmic}
\end{algorithm}

\section{Empirical Evaluation}

\subsection{Experiments on the generated maps}

\begin{figure}[t]
    \centering
    \begin{subfigure}[b]{0.2\textwidth}
        \centering
        \includegraphics[width=\textwidth]{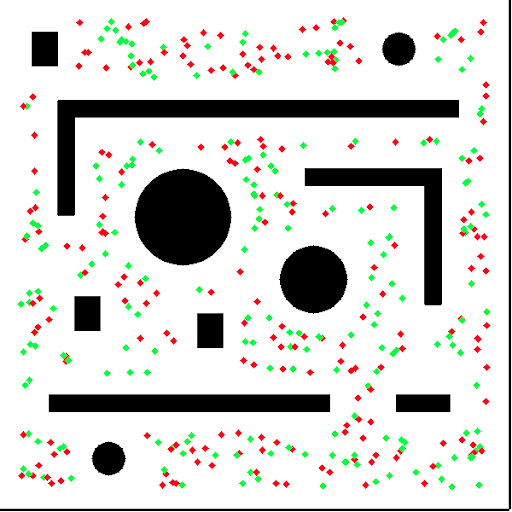}
        \caption{Map1}
        \label{fig:Map1}
    \end{subfigure}
    \begin{subfigure}[b]{0.2\textwidth}
         \centering
        \includegraphics[width=\textwidth]{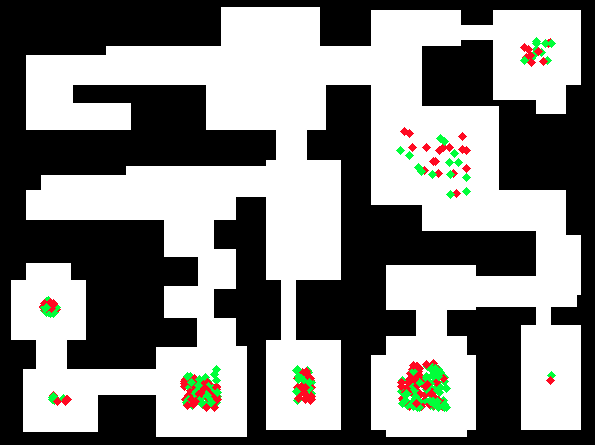}
        \caption{Map2}
        \label{fig:Map2}
    \end{subfigure}
    \begin{subfigure}[b]{0.2\textwidth}
         \centering
        \includegraphics[width=\textwidth]{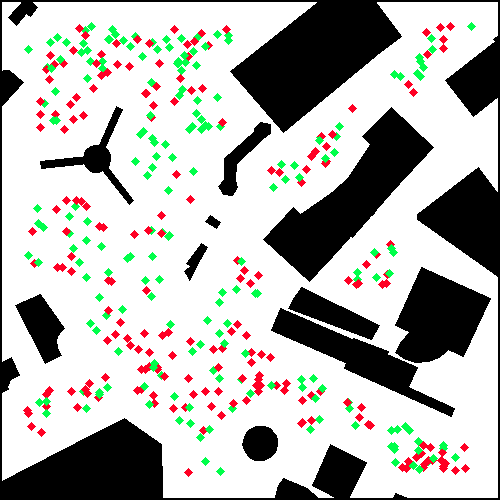}
        \caption{Map3}
        \label{fig:Map3}
    \end{subfigure}
    \begin{subfigure}[b]{0.2\textwidth}
         \centering
        \includegraphics[width=\textwidth]{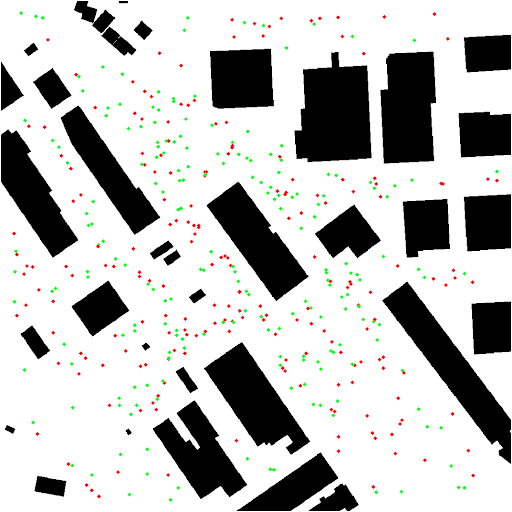}
        \caption{Map4}
        \label{fig:Map4}
    \end{subfigure}
    \caption{Maps used in our tests. \textbf{Green} and \textbf{Red} circles show the start and goal positions respectively (200 per each map).
    }
    \label{fig:environments}
\end{figure}

For the first series of the experiments, we have used four different grid-maps representing two types of the environment: indoor and outdoor -- see Fig.~\ref{fig:environments}. \texttt{Map1} and \texttt{Map2} were generated by us and have a size $600 \times 600$ and $495 \times 475$ respectively. \texttt{Map3} and \texttt{Map4} were taken from the Moving AI benchmark~\cite{Benchmarks}. They are $512 \times 512$ in size.

For each map, we generated 200 different planning tasks. For each task, we randomly generate the start and goal positions s.t. the distance from them to the nearest obstacle was at least 20 cells. Start/goal orientations were chosen randomly.

We compared our algorithm with two other methods: original GRIPS and Theta*-RRT~\cite{THETA*RRT}. All these algorithm used the geometric path constructed by Theta* planner~\cite{THETA*} to build the kinodynamic trajectory. We implemented the algorithms from scratch in C++ to favor the fair comparison. The experiments were carried out on a PC equipped with Intel Core i5-8400 CPU @ 2.80GHz.

We used the following parameters for steering function: $k_\rho=5, k_\alpha=15, k_\beta=-5$. The simulation step for steering was set to be $dt=0.1$s and the limit of steps was $300$. Assuming that each cell on a our maps is 20 cm, we simulate the movement of a car-like robot with the following parameters: maximum speed $v_{max}=2$ m/s, maximum acceleration $a_{max}=0.4$ m/s$^2$, wheel base $L=2$ m.

All the evaluated algorithms are parameterized so we conducted preliminary experiments to fine-tune them. As a result for the modified and the original versions of GRIPS we used the following parameters' values. We set the initial step size of gradient descent $\eta_{0} = 1.0$, the discount factor $\gamma = 0.8$, the number of iterations of the first phase $K = 5$ and minimal distance $d_{\min} = 1$ m. For the original GRIPS we set the maximum number of pruning rounds $L_{\max} = 50$, while for our version we used the number of iterations for the procedures \textit{Insert} $r = 1$ and \textit{Move} $K = 5$. The step of generation states $step = 1.0$ m with the maximum distance $M = 5.0$ m. 

\begin{figure}[t]
    \centering
    \begin{subfigure}[b]{0.23\textwidth}
        \includegraphics[width=\textwidth]{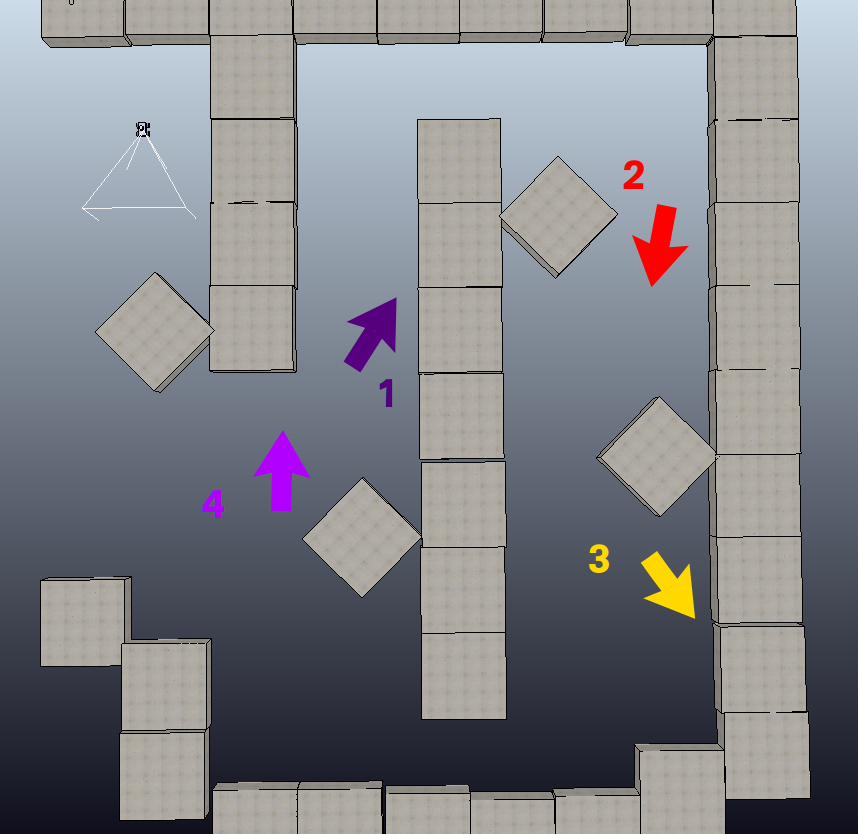}
    \end{subfigure}
    ~ 
    \begin{subfigure}[b]{0.23\textwidth}
        \includegraphics[width=\textwidth]{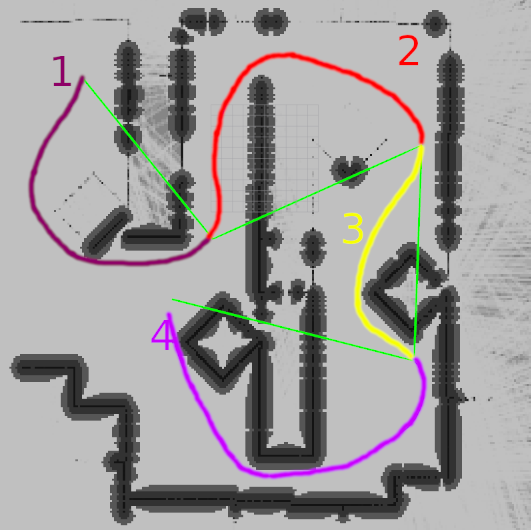}
    \end{subfigure}
    \caption{The mission and the traveled path of the robot on simulation.}
    \label{fig:MapAndTasksInSimulation}
\end{figure}

For each task, we recorded whether the evaluated algorithm managed to solve it and in case it did we documented the length of the resultant trajectory and the runtime (not taking into account time to generate geometric path). 

\begin{figure*}[t]
    \centering
    \begin{subfigure}[b]{0.32 \linewidth}
        \includegraphics[width=\linewidth]{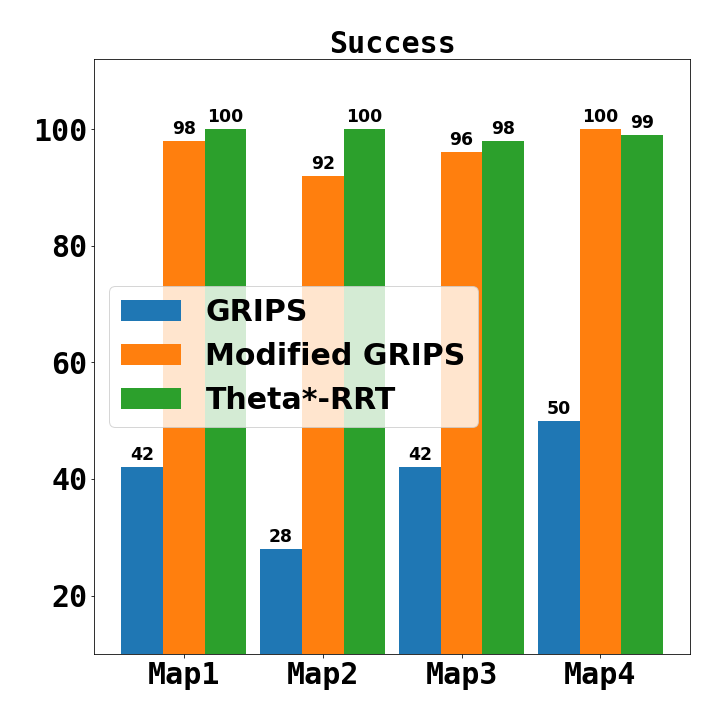}
    \end{subfigure}
    ~
    \begin{subfigure}[b]{0.32 \linewidth}
        \includegraphics[width=\linewidth]{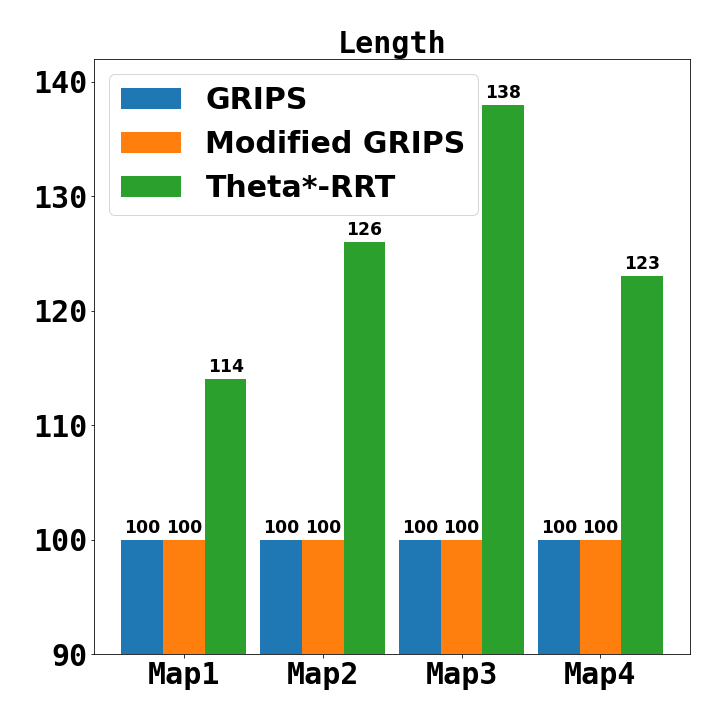}
    \end{subfigure}
    ~
    \begin{subfigure}[b]{0.32 \linewidth}
        \includegraphics[width=\linewidth]{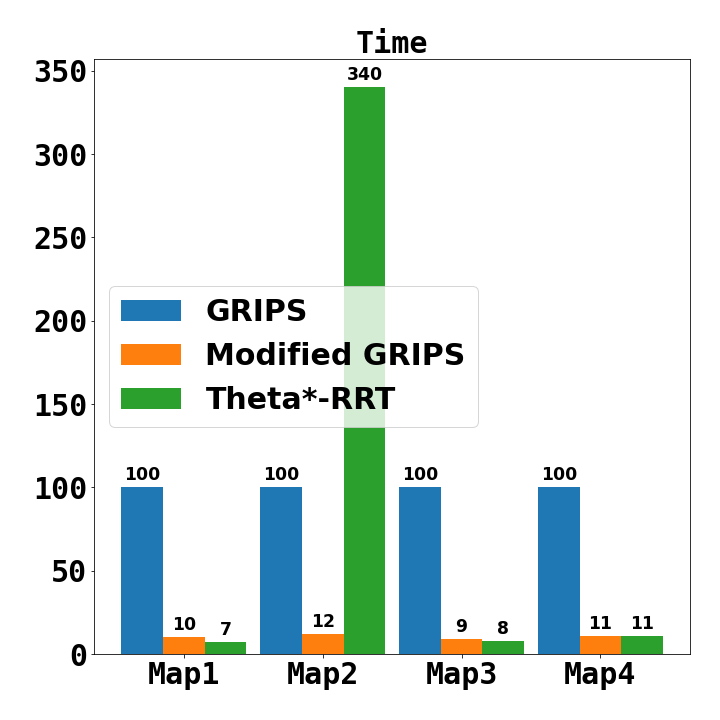}
    \end{subfigure}
    \caption{Planning results for the generated maps (success rate, average trajectory length and average algorithm's runtime).} 
    \label{fig:MetricsOnMaps}
\end{figure*}

The results are presented in Figure~\ref{fig:MetricsOnMaps}. Notably, the modified GRIPS has a much higher success rate compared to the original GRIPS. Besides the modified GRIPS is evidently faster while providing the paths of nearly the same cost. The success rate of Theta*-RRT is lightly higher compared to modified GRIPS. This is expected as Theta*-RRT is a sampling-based planner which has a much more freedom in finding the feasible trajectory. This comes at the price of longer runtimes and increase in path lengths.

\subsection{Experiments On A Simulated Robot}

To test the suggested algorithm in close-to-the-world scenarios we utilized CoppeliaSim\footnote{https://www.coppeliarobotics.com/} simulator and ready-to-go model of a wheeled robot Husky. We also utilized ROS~\cite{ROS} framework and wrapped up the planning algorithms to a ROS module which was running together with the third-party simultaneous localization and mapping software. The latter provided us with the occupancy grid and the robot's position. 

The environment used in tests is depicted on Figure~\ref{fig:MapAndTasksInSimulation}. The size of map was $120\times120$ meters which was discretized to a $600 \times 600$ grid (each cell is 20 cm). The task of the robot was to sequentially move from one landmark to the other specified by the user. Positions and orientations of the landmarks are shown on Figure~\ref{fig:MapAndTasksInSimulation} as well. When moving, robot consequently re-planned (at 10Hz) the trajectory to the current landmark from its current position. In case a re-planning attempt did not succeed robot continued moving along the previously planned trajectory. For geometric planning we used an implementation of ABIT*\cite{Gammell2020ABIT*} algorithm available from OMPL library~\cite{OMPL}. The maximum available time for a geometric planner was set to 0.07s and in practice ABIT* always managed to find geometric path under such constraints.

Table~\ref{tab:SimulationResults} shows the results. The second column says how many re-planning attempts were made when executing each of the four tasks and the last column shows how many of them were successful. As one can note the success rate is not 100\%, still the robot managed to complete the mission. This is due to the fact that even if robot occasionally failed to re-plan it used a previously planned trajectory. The runtime of the planner was always less that 0.1s which means no re-planning attempts were skipped (recall that planning module is executed at 10Hz rate).

Overall the results of the conducted experiments provide a clear evidence that the suggested algorithm can be deployed on autonomous mobile robots and plan kinodynamic trajectories in real-time.

\begin{table}[t]
\begin{center}
\begin{tabular}{p{0.05\linewidth}|p{0.20\linewidth}p{0.20\linewidth}p{0.20\linewidth}}
Task & Number of re-planning tasks & Average runtime, ms & Success rate, \% \\
\hline \hline
1 & 370 & 52 & 86\\
\hline 
2 & 469 & 54 & 83\\
\hline
3 & 262 & 46 & 89\\
\hline
4 & 445 & 64 & 75\\
\hline
\end{tabular}
\caption{Results of the experiments on simulation.}
\label{tab:SimulationResults}
\end{center}
\end{table}

\section{Conclusion}

In this paper, we proposed a modified version of GRIPS algorithm that generates a kinematically feasible trajectory for a car-like robot provided with a geometric path and a steering function. The modified algorithm is notably faster and has a much higher success rate than a predecessor.  We used the described method as a part of the control system of a simulated robot and showed that the latter was able to solve different path-planning tasks under a tight computational budget (i.e. produce a plan each 0.1s).

\bibliographystyle{ieeetr}
\bibliography{root}

\end{document}